\documentclass[runningheads,a4paper]{llncs}
\usepackage{graphicx}
\usepackage{subfigure}

\begin{document}

\title{Video-based computer aided arthroscopy for patient specific reconstruction of the Anterior Cruciate Ligament\thanks{The authors thank the Portuguese Science Foundation
and COMPETE2020 program for generous funding through
project VisArthro (ref.: PTDC/EEIAUT/3024/2014). This
paper was also funded by the European Union's Horizon 2020 research and innovation programme under grant
agreement No 766850.}}

\titlerunning{Computer-aided arthroscopy for ACL reconstruction}  
%

\author{Carolina Raposo\inst{1,2} \and
Crist\'{o}v\~{a}o Sousa\inst{2} \and
Luis Ribeiro\inst{2} \and
Rui Melo\inst{2} \and
Jo\~{a}o P. Barreto\inst{1,2} \and
Jo\~{a}o Oliveira\inst{3} \and
Pedro Marques\inst{3} \and
Fernando Fonseca\inst{3}}
\authorrunning{C. Raposo \emph{et al.}} 
\institute{Institute of Systems and Robotics, University of Coimbra, Portugal 
\and
Perceive3D, Coimbra, Portugal 
\and
Coimbra Hospital and Universitary Centre, Faculty of Medicine, Coimbra, Portugal} 

\maketitle              

\begin{abstract}
The Anterior Cruciate Ligament (ACL) tear is a common medical condition that is treated using arthroscopy by pulling a tissue graft through a tunnel opened with a drill. The correct anatomical position and orientation of this tunnel is crucial for knee stability, and drilling an adequate bone tunnel is the most technically challenging part of the procedure. This paper presents, for the first time, a guidance system based solely on intra-operative video for guiding the drilling of the tunnel. Our solution uses small, easily recognizable visual markers that are attached to the bone and tools for estimating their relative pose. A recent registration algorithm is employed for aligning a pre-operative image of the patient's anatomy with a set of contours reconstructed by touching the bone surface with an instrumented tool. Experimental validation using ex-vivo data shows that the method enables the accurate registration of the pre-operative model with the bone, providing useful information for guiding the surgeon during the medical procedure.
\end{abstract}
\section{Introduction}
\begin{figure}
\centering
\subfigure[Reconstruction of 3D contours]{\includegraphics[width = 0.43\linewidth]{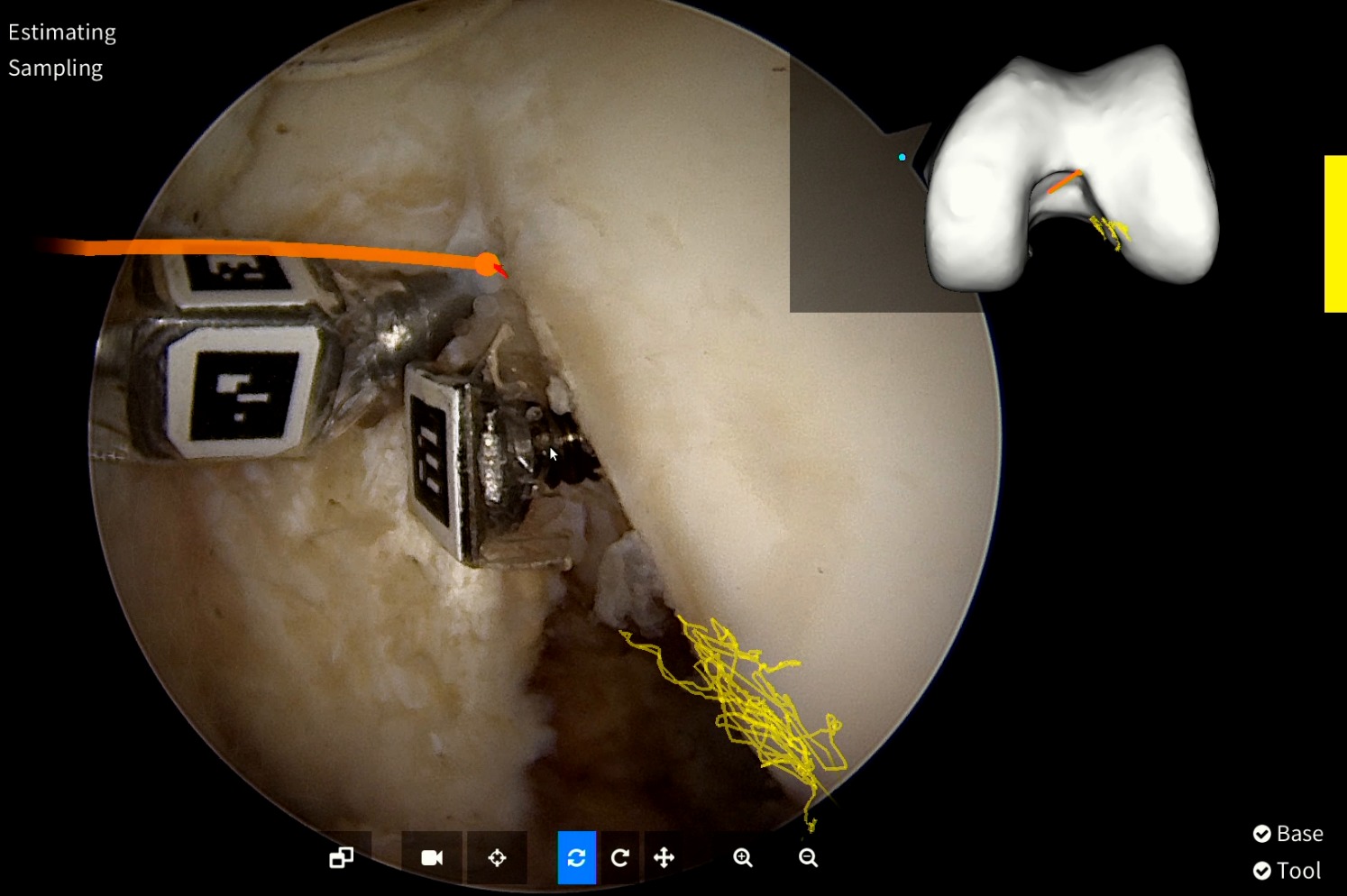}\label{fig:intro1}} \qquad
\subfigure[Result of the 3D alignment]{\includegraphics[width = 0.43\linewidth]{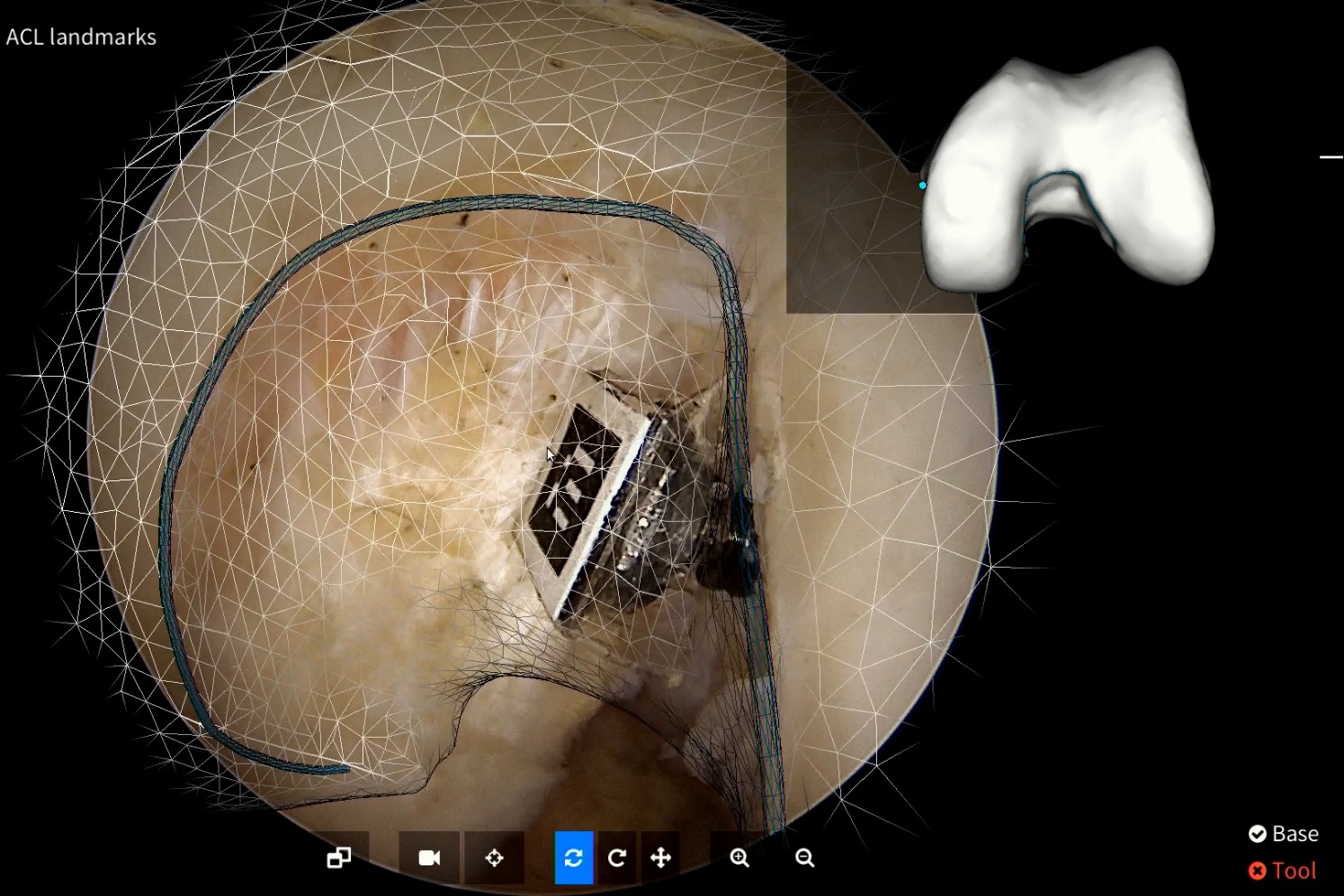}\label{fig:intro2}}
\caption{\subref{fig:intro1} 3D digitalisation of the bone surface: the surgeon performs a random walk on the intercondylar region using a touch-probe instrumented with a visual marker with the objective of reconstructing 3D curves on the bone surface. \subref{fig:intro2} Overlay of the pre-operative MRI with highlight of intercondylar arch: The reconstructed 3D curves are used to register the pre-operative MRI with the patient anatomy.}
\label{fig:intro}
\end{figure}
Arthroscopy is a modality of orthopeadic surgery for treatment of damaged joints in which instruments and endoscopic camera (the arthroscope) are inserted into the articular cavity through small incisions (the surgical ports). Since arthroscopy largely preserves the integrity of the articulation, it is beneficial to the patient in terms of reduction of trauma, risk of infection and recovery time~\cite{Treuting}. However, arthroscopic approaches are more difficult to execute than the open surgery alternatives because of the indirect visualization and limited manoeuvrability inside the joint, with novices having to undergo a long training period~\cite{Nawabi} and experts often making mistakes with clinical consequences~\cite{Samitier}. 

The reconstruction of the ACL illustrates well the aforementioned situation. The ACL rupture is a common medical condition with more than 200 000 annual cases in the USA alone~\cite{Samitier}. The standard way of treatment is arthroscopic reconstruction where the torn ligament is replaced by a tissue graft that is pulled into the knee joint through tunnels opened with a drill in both femur and tibia~\cite{Brown}. Opening these tunnels in an anatomically correct position is crucial for knee stability and patient satisfaction, with the ideal graft being placed in the exact same position of the original ligament to maximize proprioception~\cite{Barrett}. Unfortunately, ligament position varies significantly across individuals and substantial effort has been done to model variance and provide anatomic references to be used during surgery~\cite{Forsythe}. However, correct tunnel placement is still a matter of experience with success rates varying broadly between low and high volume surgeons~\cite{Samitier}. Some studies reveal levels of satisfaction of only 75$\%$ with an incidence of revision surgeries of 10 to 15$\%$, half of which caused by deficient technical execution~\cite{Samitier}.

This is a scenario where Computer-Aided Surgery (CAS) can have an impact. There are two types of navigation systems reported in literature: the ones that use intra-operative fluoroscopy~\cite{Brown}, and the ones that rely in optical tracking to register a pre-operative CT/MRI or perform 3D bone morphing~\cite{Kim}. Despite being available for several years the market, penetration of these systems is residual because of their inaccuracy and inconvenience~\cite{Kim}. The added value of fluoroscopy based systems does not compensate the risk of radiation overdose, while optical tracking systems require additional incisions to attach markers which hinders acceptance because the purpose of arthroscopy is to minimize incisions. The ideal solution for Computer-Aided Arthroscopy (CAA) should essentially rely in processing the already existing intra-operative video. This would avoid the above mentioned issues and promote cost efficiency by not requiring additional capital equipment. Despite the intense research in CAS using endoscopic video~\cite{Maier}, arthroscopic sequences are specially challenging because of poor texture, existence of deformable tissues, complex illumination, and very close range acquisition. In addition, the camera is hand-held, the lens scope rotates, the procedure is performed in wet medium and the surgeon often switches camera port. Our attempts of using visual SLAM pipelines reported to work in laparoscopy~\cite{Mahmoud} were unfruitful and revealed the need of additional visual aids to accomplish the robustness required for real clinical uptake.    

This article describes the first video-based system for CAA, where visual information is used to register a pre-operative CT/MRI with the patient anatomy such that tunnels can be opened in the position of the original ligament (patient specific surgery). The concept relates with previous works in CAS for laparoscopy that visually track a planar pattern engraved in a projector to determine its 3D pose~\cite{Edgcumbe}. We propose to attach similar fiducial markers to both anatomy and instruments and use the moving arthroscope to estimate the relative rigid displacements at each frame time instant. The scheme enables to perform accurate 3D reconstruction of the bone surface with a touch-probe (Fig.~\ref{fig:intro1}) that is used to accomplish registration of the pre-operative 3D model or plan (Fig.~\ref{fig:intro2}). The marker of the femur (or tibia) works as the world reference frame where all 3D information is stored, which enables to quickly resume navigation after switching camera port and overlay guidance information in images using augmented reality techniques. The paper describes the main modules of the real-time software pipeline and reports results in both synthetic and real \textit{ex-vivo} experiments.

\section{Video-based computer-aided arthroscopy}
\begin{figure}
\centering
\subfigure[Pose estimation]{\includegraphics[width = 0.32\linewidth]{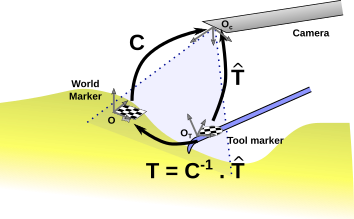}\label{fig:step1}} \hfill
\subfigure[Contour reconstruction]{\includegraphics[width = 0.32\linewidth]{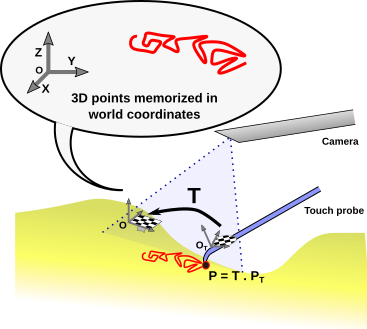}\label{fig:step2}} \hfill
\subfigure[Registration $\&$ guidance]{\includegraphics[width = 0.32\linewidth]{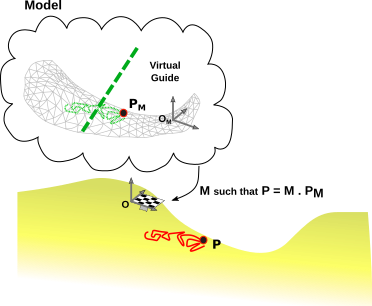}\label{fig:step3}} \hfill
\caption{Key steps of the proposed approach: \subref{fig:step1} 3D pose estimation inside the articular joint, \subref{fig:step2} 3D reconstruction of points and contours on bone surface and \subref{fig:step3} 3D registration and guidance.}
\label{fig:funcs}
\end{figure}

This section overviews the proposed concept for CAA that uses the intra-operative arthroscopic video, together with planar visual markers attached to instruments and anatomy, to perform tracking and 3D pose estimation inside the articular joint. As discussed, applying conventional SLAM/SfM algorithms~\cite{Mahmoud} to arthroscopic images is extremely challenging and, in order to circumvent the difficulties, we propose to use small planar fiducial markers that can be easily detected in images and whose pose can be estimated using homography factorization~\cite{Ma,Belhaoua}. These visual aids enable to achieve the robustness and accuracy required for deployment in real arthroscopic scenario that otherwise would be impossible. The key steps of the approach are the illustrated in Fig.~\ref{fig:funcs} and described next.

\textbf{The anatomy marker WM}:
The surgeon starts by rigidly attaching a screw-like object with a flat head that has an engraved known 4mm-side square pattern. We will refer to this screw as the World Marker (WM) because the local reference frame of its pattern will define the coordinate system with respect to which all 3D information is described. The WM can be placed in an arbitrary position in the intercondylar surface, provided that it can be easily seen by the arthroscope during the procedure. The placement of the marker is accomplished using a custom made tool that can be seen in the accompanying video.

\textbf{3D pose estimation inside the articular joint}:
The 3D pose $\mathsf{C}$ of the WM in camera coordinates can be determined at each time instant by detecting the WM in the image, estimating the plane-to-image homography from the 4 detected corners of its pattern and decomposing the homography to obtain the rigid transformation~\cite{Ma,Belhaoua}. Consider a touch probe that is also instrumented with another planar pattern that can be visually read. 
Using a similar method, it is possible to detect and identify the tool marker (TM) in the image and compute its 3D pose $\mathsf{\hat{T}}$ with respect to the camera. This allows the pose $\mathsf{T}$ of the TM in WM coordinates to be determined in a straightforward manner by $\mathsf{T} = \mathsf{C}^{-1}\mathsf{\hat{T}}$. 

\textbf{3D reconstruction of points and contours on bone surface}:
The location of the tip of the touch probe in the local TM reference frame is known, meaning that its location w.r.t. the WM can be determined using $\mathsf{T}$.
A point on the surface can be determined by touching it with the touch probe. A curve and/or sparse bone surface reconstruction can be accomplished in a similar manner by performing a random walk.

\textbf{3D registration and guidance}:
The 3D reconstruction results are used to register the 3D pre-operative model, enabling to overlay the plan with the anatomy. We will discuss in more detail how this registration is accomplished in the next section.
The tunnel can be opened using a drill guide instrumented with a distinct visual marker and whose symmetry axis's location is known in the marker's reference frame. This way, the location of the drill guide w.r.t. the pre-operative plan is known, providing real-time guidance to the surgeon, as shown in the accompanying video.

This paper will not detail the guidance process as it is more a matter of application as soon as registration is accomplished. Also, it will solely refer to the placement of femoral tunnel, with the placement of tibial tunnel being similar.

\section{Surgical workflow and algorithmic modules}
\begin{figure}
\centering
\subfigure[Surgical workflow]{\includegraphics[width = 0.42\linewidth]{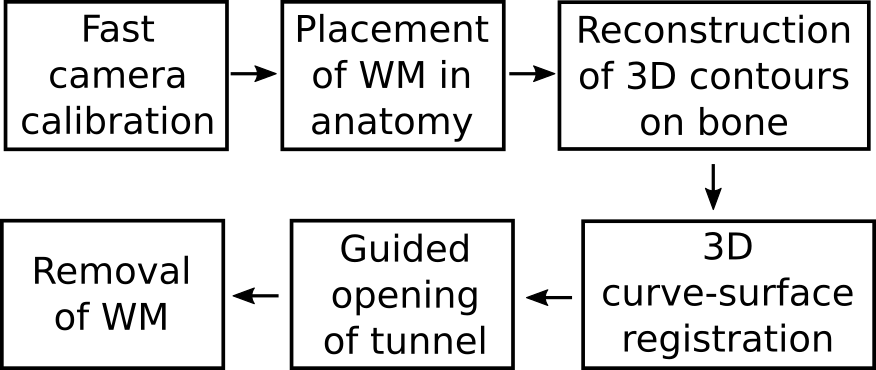}\label{fig:steps}} \hfill
\subfigure[Registration process]{\includegraphics[width = 0.50\linewidth]{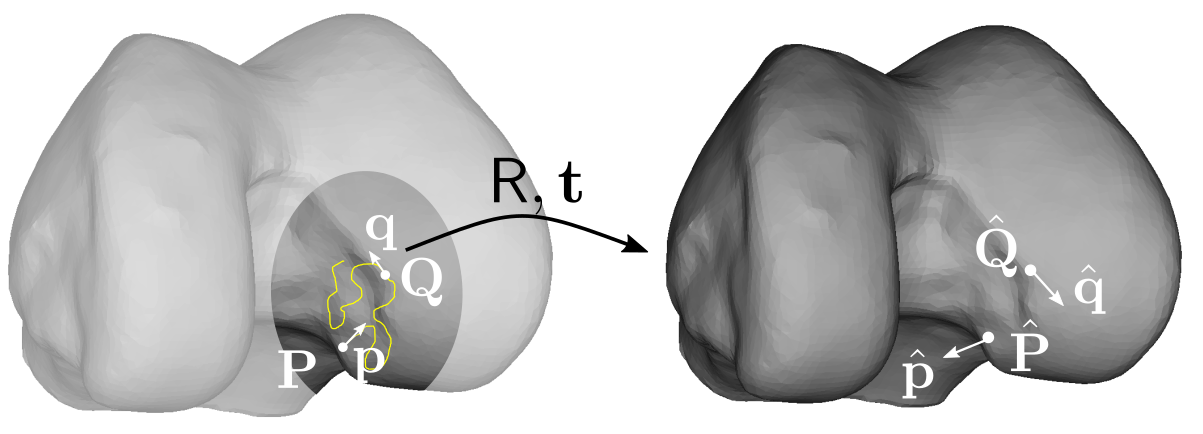}\label{fig:registscheme}} 
\caption{\subref{fig:steps} Different steps of the surgical workflow and \subref{fig:registscheme} the rigid transformation $\mathsf{R}, \mathbf{t}$ is determined by searching for pairs of points $\mathbf{P}, \mathbf{Q}$ with tangents $\mathbf{p},\mathbf{q}$ on the curve side that are a match with pairs of points $\mathbf{\hat{P}}, \mathbf{\hat{Q}}$ with normals $\mathbf{\hat{p}}, \mathbf{\hat{q}}$ on the surface side.}
\label{fig:3}
\end{figure}

The steps of the complete surgical workflow are given in Fig.~\ref{fig:steps}. An initial camera calibration using 3 images of a checkerboard pattern with the lens scope at different rotation angles is performed. Then, the world marker is rigidly attached to the anatomy and 3D points on the bone surface are reconstructed by scratching it with an instrumented touch probe. While the points are being reconstructed and using the pre-operative model, the system performs an on-the-fly registration that allows the drilling of the tunnel to be guided. Guidance information is given using augmented reality, by overlaying the pre-operative plan with the anatomy in real time, and using virtual reality, by continuously showing the location of the drill guide in the model reference frame. As a final step, the WM must be removed.
Details are given below.

\textbf{Calibration in the OR}:
Since the camera has exchangeable optics, calibration must be made in the OR. In addition, the lens scope rotates during the procedure, meaning that intrinsics must be adapted on the fly for greater accuracy. This is accomplished using an implementation of the method described in~\cite{Melo}. Calibration is done by collecting 3 images rotating the scope to determine intrinsics, radial distortion and center of rotation. For facilitating the process, acquisition is carried in dry environment and adaptation for wet is performed by multiplying the focal length by the ratio of the refractive indices~\cite{Lavest}.

\textbf{Marker detection and pose estimation}: There are several publicly available libraries for augmented reality that implement the process of detection, identification and pose estimation of square markers. We opted for the ARAM library~\cite{Belhaoua} and, for better accuracy, also used a photogeometric refinement step as described in~\cite{Mei} with the extension to accommodate radial distortion as in~\cite{Lourenco}, making possible the accommodation of variable zoom in a future version.

\textbf{Registration}: Registration is accomplished using the method in~\cite{Raposo} that uses a pair of points $\mathbf{P}, \mathbf{Q}$ with tangents $\mathbf{p},\mathbf{q}$ from the curve that matches a pair of points $\mathbf{\hat{P}}, \mathbf{\hat{Q}}$ with normals $\mathbf{\hat{p}}, \mathbf{\hat{q}}$ on the surface for determining the alignment transformation (Fig.~\ref{fig:registscheme}). The search for correspondences is performed using a set of conditions that also depend on the differential information. Global registration is accomplished using an hypothesise-and-test framework.

\textbf{Instruments and Hardware setup}:
The software runs in a PC that is connected in-between camera tower and display. The PC is equipped with a frame grabber Datapath Limited DGC167 in an Intel Core i7 4790 and a GPU NVIDIA GeForce GTX950 that was able to run the pipeline in HD format at 60fps with latency of 3 frames.
In addition, we built the markers, custom screw removal tool, touch probe and drill guide that can be seen in the video.

\begin{figure}[t]
\centering
\includegraphics[width = 0.85\linewidth]{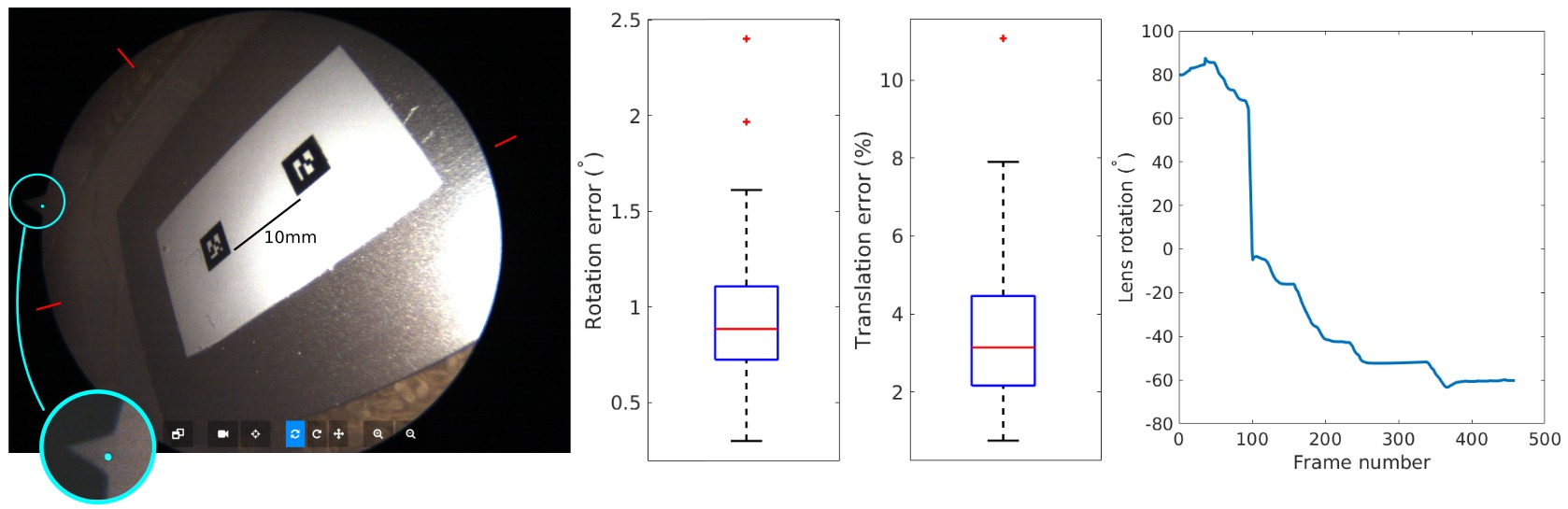}
\caption{Experiment on lens rotation in wet environment.}
\label{fig:rotlens}
\end{figure}
\section{Experiments}
This section reports experiments that assess the performance of two key features of the presented system: the compensation of the camera's intrinsics according to the rotation of the lens and the registration of the pre-operative model with the patient's anatomy. Tests on laboratory and using \textit{ex-vivo} data are performed.
\subsection{Lens rotation}
This experiment serves to assess the accuracy of the algorithm for compensating the camera's intrinsics according to the scope's rotation. We performed an initial camera calibration using 3 images of a calibration grid with the scope at 3 different rotation angles, which are represented with red lines in the image on the left of Fig.~\ref{fig:rotlens}. We then acquired a 500-frame video sequence of a ruler with two 2.89mm-side square markers 10mm apart in wet environment. The rotation of the scope performed during the acquisition of the video is quantified in the plot on the right of Fig.~\ref{fig:rotlens} that shows that the total amount of rotation was more than 140$^\circ$. The lens mark, shown in greater detail in  Fig.~\ref{fig:rotlens}, is detected in each frame for compensating the intrinsics. The accuracy of the method is evaluated by computing the relative pose between the two markers in each frame and comparing it with the ground truth pose. The low rotation and translation errors show that the algorithm properly handles lens rotation.
\subsection{3D Registration}
\begin{figure}
\centering
\subfigure[Landmarks]{\includegraphics[width = 0.31\linewidth]{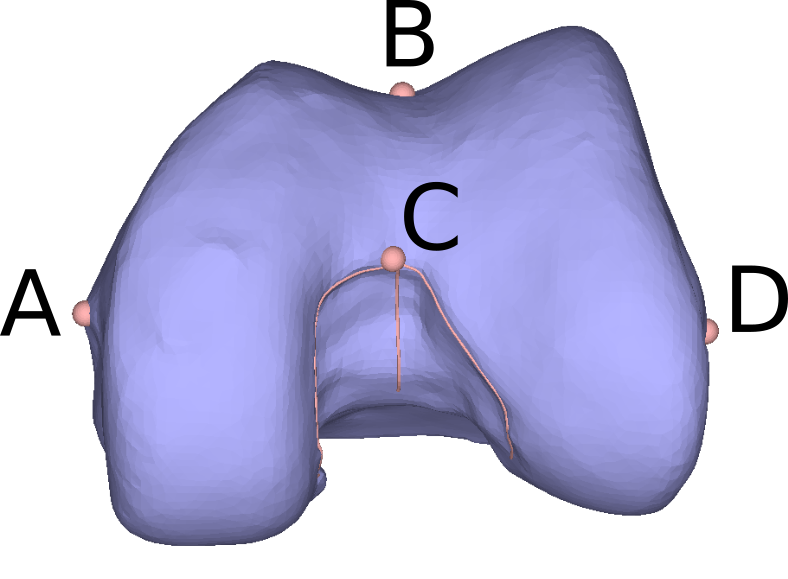}\label{fig:cad1}} \hfill
\subfigure[Experimental setup]{\includegraphics[width = 0.28\linewidth]{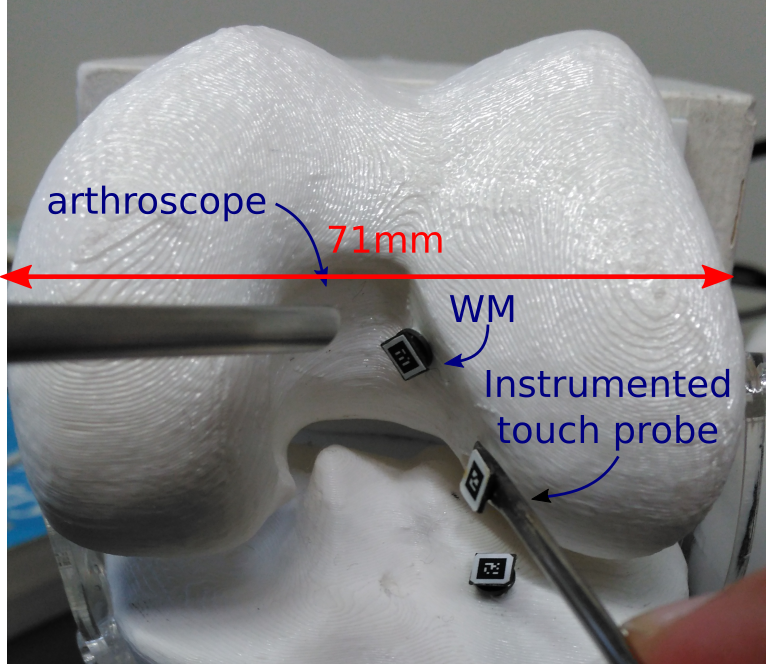}\label{fig:cad2}} \hfill
\subfigure[Registration results]{\includegraphics[width = 0.34\linewidth]{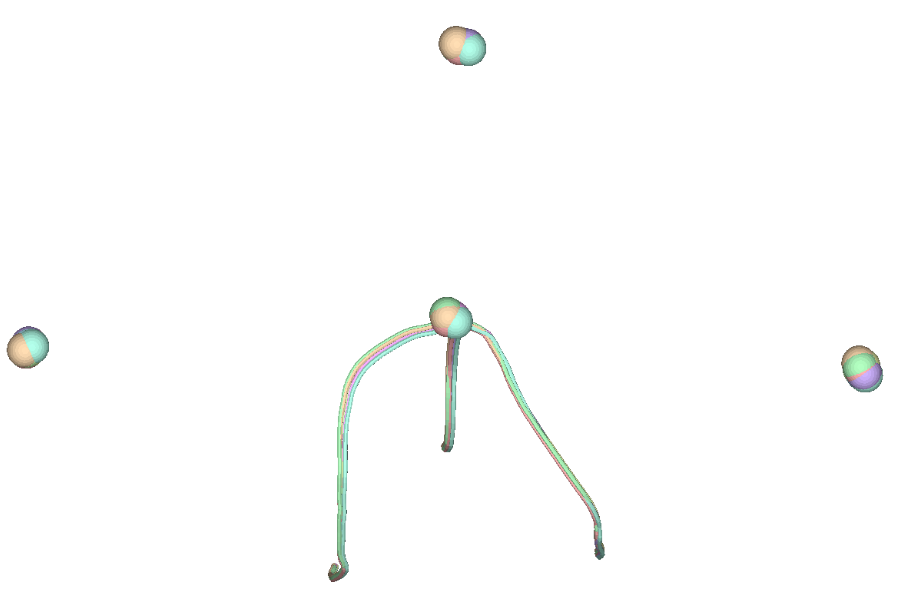}\label{fig:cad3}}

\subfigure[Experimental setup]{\includegraphics[width = 0.280\linewidth]{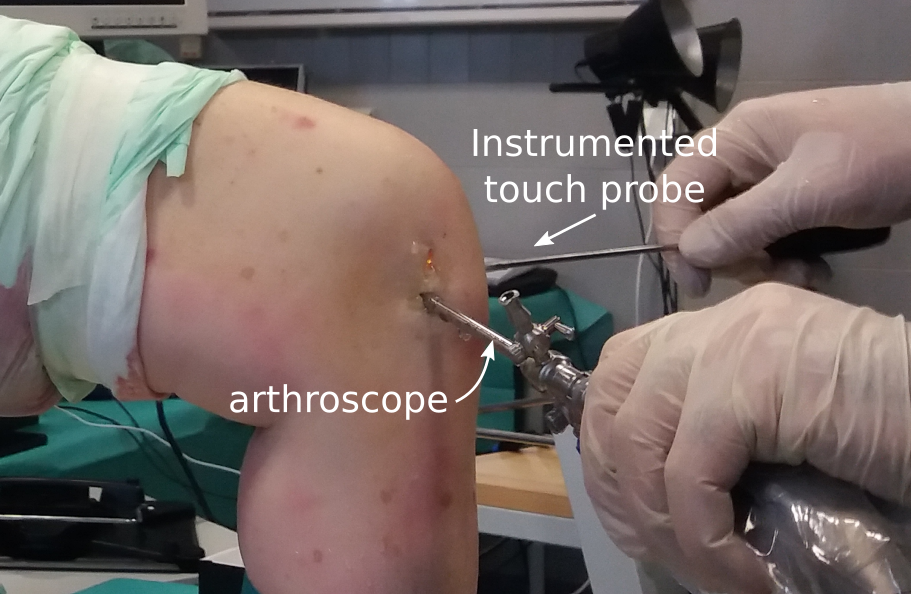}\label{fig:cad4}} \hfill
\subfigure[Registration results]{\includegraphics[width = 0.30\linewidth]{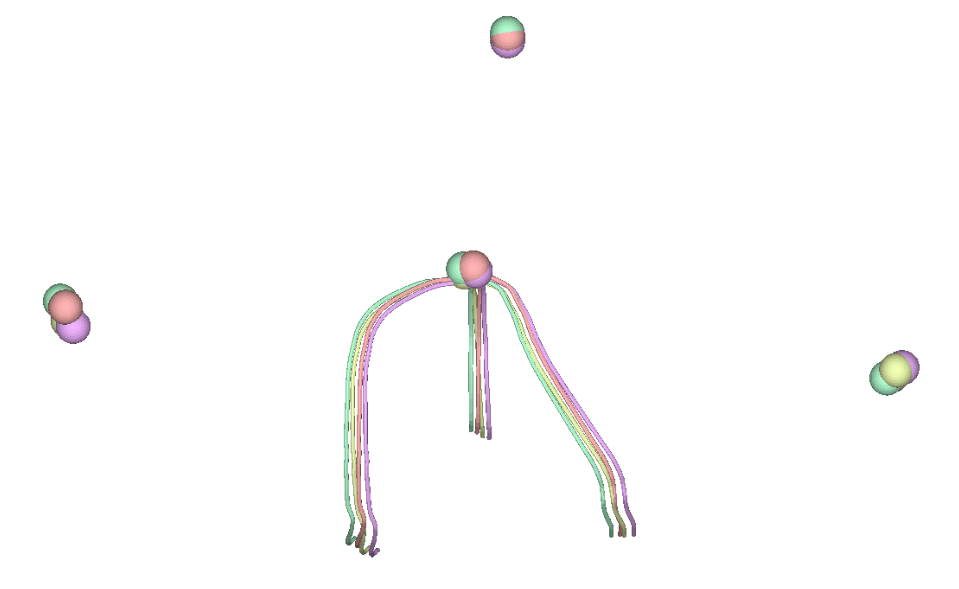}\label{fig:cad5}} \hfill
\subfigure[Quantitative eval.]{\includegraphics[width = 0.27\linewidth]{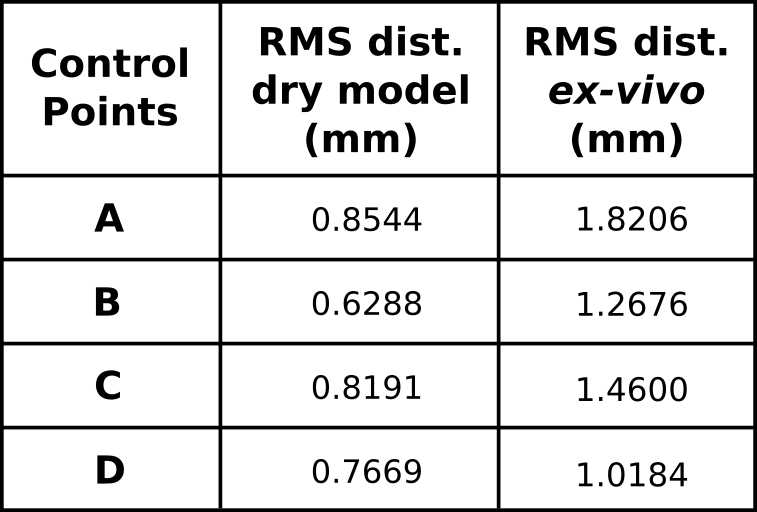}\label{fig:cad6}}
\caption{Analysis of performance of the registration algorithm in two experiments: one in the laboratory using a dry knee model and another using \textit{ex-vivo} data.}
\label{fig:cadsynth}
\end{figure}
The first test regarding the registration method was performed on a dry model and consisted in reconstructing 10 different sets of curves by scratching the rear surface of the lateral condyle with an instrumented touch probe, and registering them with the virtual model shown in Fig.~\ref{fig:cad1}, providing 10 different rigid transformations. A qualitative assessment (Fig.~\ref{fig:cad3}) of the registration accuracy is provided by representing the anatomical landmarks and the control points of Fig.~\ref{fig:cad1} in WM coordinates using the obtained transformations. The centroid of each point cloud obtained by transforming the control points is computed and the RMS distance between each transformed point and the corresponding centroid is computed and shown in Fig.~\ref{fig:cad6}, providing a quantitative assessment of the registration accuracy. Results show that all the trials provided very similar results, with the landmarks and control points being almost perfectly aligned in Fig.~\ref{fig:cad3} and all RMS distances being below 0.9mm, despite the control points belonging to regions that are very distant from the reconstructed area.

The second experiment was performed on \textit{ex-vivo} data and followed a similar strategy as the one on the dry model, having the difference that the total number of trials was 4. Fig.~\ref{fig:cad4} illustrates the setup of the \textit{ex-vivo} experiment and Figs~\ref{fig:cad5} and \ref{fig:cad6} show the qualitative and qualitative analyses of the obtained result. Results show a slight degradation in accuracy w.r.t. the dry model test, which is expected since the latter is a more controlled environment. However, the obtained accuracy is very satisfactory, with the RMS distances of all control points being below 2mm. This experiment demonstrates that our proposed system is very accurate in aligning the anatomy with a pre-operative model of the bone, enabling a reliable guidance of the ACL reconstruction procedure.
\section{Conclusions}
This paper presents the first video-based navigation system for ACL reconstruction. The software is able to handle unconstrained lens rotation and register pre-operative 3D models with the patient's anatomy with high accuracy, as demonstrated by the experiments performed both on a dry model and using \textit{ex-vivo} data. This allows the complete medical procedure to be guided, leading not only to a significant decrease in the learning curve but also to the avoidance of technical mistakes. 
As future work, we will be targeting other procedures that might benefit from navigation such as resection of Femuro Acetabular Impingement during hip arthroscopy.
%
%

\bibliographystyle{splncs03}
\bibliography{sample}

\end{document}